\pdfoutput=1

\documentclass[11pt]{article}

\usepackage{acl}
\usepackage{times}
\usepackage{latexsym}

\usepackage[T1]{fontenc}

\usepackage[utf8]{inputenc}

\usepackage{microtype}

\usepackage{microtype}
\usepackage{hyperref}
\usepackage{url}
\usepackage{booktabs}
\usepackage{multirow}
\usepackage{makecell}
\usepackage{soul}
\usepackage{enumitem}

\usepackage{amsthm}
\theoremstyle{plain}

\theoremstyle{claim}

\theoremstyle{fact}

\theoremstyle{definition}
\newtheorem{definition}{Definition}
\theoremstyle{lemma}

\theoremstyle{phenomenon}
\newtheorem{phenomenon}{Phenomenon}
\usepackage{wrapfig,lipsum,booktabs}
\theoremstyle{proper}

\theoremstyle{remark}
\usepackage{wrapfig}
\usepackage{algorithm}
\usepackage{algpseudocode}
\usepackage{graphicx}
\usepackage{xspace}

\newcommand{\modelPrefix}{AraLLaMA}
\newcommand{\model}{\modelPrefix\xspace}
\newcommand{\modelLarge}{\modelPrefix-13B\xspace}
\newcommand{\modelSmall}{\modelPrefix-7B\xspace}
\newcommand{\modelLargeBase}{\modelPrefix-13B-base\xspace}
\newcommand{\modelSmallBase}{\modelPrefix-7B-base\xspace}
\newcommand{\modelLargeChat}{\modelPrefix-13B-chat\xspace}
\newcommand{\modelSmallChat}{\modelPrefix-7B-chat\xspace}

\newcommand{\jianqing}[1]{{ \color{black} #1}}
\newcommand{\juhao}[1]{{ \color{black} #1}}
\usepackage{setspace}

\title{Second Language (Arabic) Acquisition of LLMs via \\Progressive Vocabulary Expansion }

\author{\textbf{Jianqing Zhu}$^\dagger$$^1$,
        \textbf{Huang Huang}$^\dagger$$^2$,
        \textbf{Zhihang Lin}$^\dagger$$^4$,
        \textbf{Juhao Liang}$^\dagger$$^{3,4}$,
        \textbf{Zhengyang Tang}$^\dagger$$^{3,4}$ \\
        \textbf{Khalid Almubarak}$^{6}$,
        \textbf{Abdulmohsen Alharthik}$^{1}$,
        \textbf{Bang An}$^{1}$,
        \textbf{Juncai He}$^{1}$,
        \textbf{Xiangbo Wu}$^{3}$,\\ 
        \textbf{Fei Yu}$^{3,4}$,
        \textbf{Junying Chen}$^{3,4}$,
        \textbf{Zhuoheng Ma}$^{4}$,
        \textbf{Yuhao Du}$^{4}$,
        \textbf{He Zhang}$^{4}$,
        \textbf{Saied Alshahrani}$^{7}$, \\
        \textbf{Emad A. Alghamdi}$^{5}$, 
        \textbf{Lian Zhang}$^{2}$,
        \textbf{Ruoyu Sun}$^{3,4}$,
        \textbf{Haizhou Li}$^{3,4}$,\\
        \textbf{Benyou Wang}\thanks{corresponding author: \texttt{wangbenyou@cuhk.edu.cn}. \\ $^\dagger$ These authors contributed equally to this work.}$^{3,4}$,
        \textbf{Jinchao Xu}$^{1}$\\
        $^1$ King Abdullah University of Science and Technology 
        $^2$ Shenzhen International Center \\for Industrial and Applied Mathematics, Shenzhen Research Institute of Big Data\\
        $^3$ Shenzhen Research Institute of Big Data
        $^4$ The Chinese University of Hong Kong \\
        $^5$ King Abdulaziz University
        $^6$ Prince Sattam bin Abdulaziz University
        $^7$ University of Bisha\\
        \texttt{}
        }

\begin{document}

\maketitle
\vspace*{10pt}
\begin{abstract}
This paper addresses the critical need for democratizing large language models (LLM) in the Arab world, a region that has seen slower progress in developing models comparable to state-of-the-art offerings like GPT-4 or GPT-3.5, due to a predominant focus on mainstream languages (e.g., English and Chinese). 
One practical objective for Arabic LLMs is to utilize Arabic-specific vocabulary in the tokenizer to accelerate decoding. However, using a different vocabulary often leads to degradation of the model's learned knowledge, since many words become out-of-vocabulary (OOV) at the beginning of training.
Inspired by the vocabulary learning during Second Language (Arabic) Acquisition for humans,  the released \model employs progressive vocabulary expansion, which is implemented by a modified BPE algorithm that progressively extends the Arabic subwords in its dynamic vocabulary during training, thereby balancing the OOV ratio at every stage. 
The ablation study demonstrated the effectiveness of Progressive Vocabulary Expansion.
Moreover, \model achieves decent performance comparable to the best Arabic LLMs across a variety of Arabic benchmarks. 
Our model weights are available at: {\url{https://github.com/FreedomIntelligence/AraLLaMa}}. 
\end{abstract}

\section{Introduction}

In the evolving landscape of large language models (LLMs), the predominant focus has been on English and Chinese. This focus has left other linguistic communities, notably the Arab world, with slower progress in developing comparable models. 
Within the Arab world \footnote{The Arab World comprises a large group of countries, mainly located in Western Asia and Northern Africa.}, the development\\
\\
of models such as Jais~\cite{sengupta2023jais} and AceGPT~\cite{huang2024acegpt} marks a significant step forward, yet these models do not rival the capabilities of state-of-the-art models like GPT-4~\cite{achiam2023gpt} or even GPT-3.5. 
In line with democratization~\cite{touvron2023llama1, touvron2023llama}, our development of Arabic LLMs focuses on language adaptation settings that utilize existing standard LLM architectures (like LLaMA~\cite{touvron2023llama}) and well-trained weights, thus saving computing resources and ensuring compatibility.

A primary challenge in adapting English-centric LLMs to other languages lies in vocabulary expansion~\cite{touvron2023llama,cui2023efficient,huang2024acegpt,zhao2024llama}. For instance, AceGPT exhibits slower decoding speeds when processing Arabic, which may be attributed to limitations in its vocabulary adaptation mechanisms.
It decodes Arabic words into sequences of alphabetical letters rather than at a more efficient granularity, such as Arabic subwords. This inefficiency significantly limits its broader applicability, despite its performance being nearly on par with GPT-3.5 in some benchmarks. A key concern related to vocabulary expansion is the risk that abrupt increases may result in a high incidence of out-of-vocabulary (OOV) tokens—units absent from the model's established vocabulary. Such a surge in OOV words can compromise the linguistic knowledge embedded within the core models. Addressing this issue requires a considerable volume of pre-training data to restore and maintain the model's linguistic capabilities effectively.

The core philosophy behind \model is inspired by the process of vocabulary learning in human Second Language Acquisition, emphasizing that individuals typically expand their vocabulary gradually through incremental learning, rather than through instantaneous acquisition.
\model progressively extends the Arabic subwords in its vocabulary during pre-training, effectively reducing the ratio of OOV words at every stage. \model, based on the initialization of LLaMA2~\cite{touvron2023llama}, not only retains the foundational knowledge of LLaMA2, but also enables effective cross-lingual transfer from English to Arabic. Ablation on  TinyLLaMA~\cite{zhang2024tinyllama}  demonstrated the effectiveness of the proposed progressive vocabulary expansion, see Section~\ref{sec:ablation}.

Followed by extensive instruction tuning, \model achieves decent performance comparable to the best Arabic LLMs across various Arabic benchmarks. The contributions of this work are three-fold: 
1) We introduce progressive vocabulary expansion, utilizing a modified byte pair encoding (BPE) algorithm inspired by human Second Language Acquisition, and demonstrate its effectiveness.
2) We present \model, a pioneering open-source Arabic Large Language Model that decodes Arabic texts three times faster than its predecessor~\cite{huang2024acegpt} while delivering superior performance.
3) We provide the community with access to the complete data processing pipeline, pre-training/fine-tuning data, and model weights. \model is compatible with the most popular LLM architecture (i.e., LLaMA) and can be seamlessly integrated into most LLM applications.

\section{Motivation: Second Language Acquisition for Humans and LLMs}

\subsection{Cognitively-inspired Motivation: Second Language Acquisition for Humans}

\begin{figure*}[t]
\centering
\includegraphics[width=\textwidth]{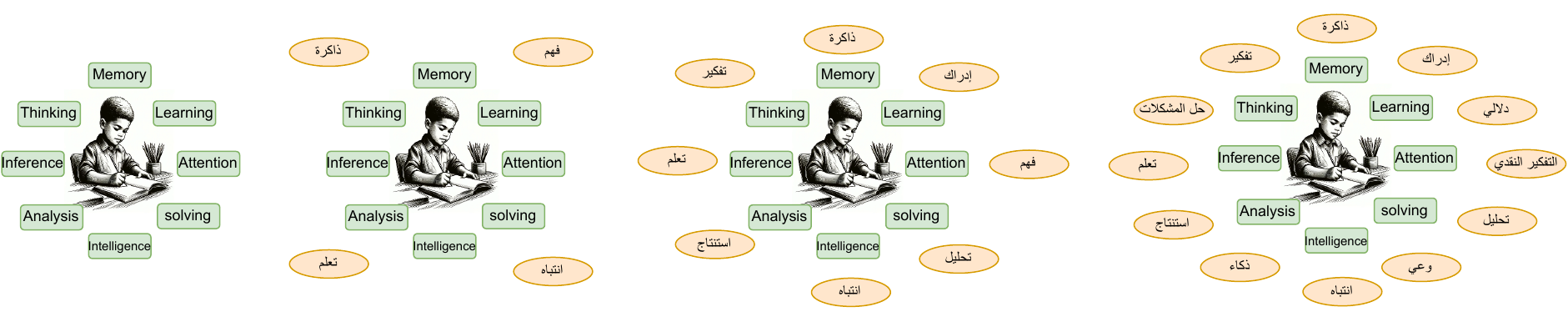}
\caption{Second Language Acquisition for human, an English-speaking child's journey to Arabic fluency, from basic vocabulary to cultural roficiency}
\label{fig:boy_learn_arabic}
\end{figure*}
\begin{definition}
    \textbf{Second Language Acquisition~(SLA)} refers to the process by which people learn a language other than their native language~\cite{krashen1981second}. SLA can occur through formal instruction in an educational setting or informally through social interaction and exposure to the language in natural settings. 
\end{definition}

In learning a second language (L2), learners pass through several developmental stages as they gain proficiency in L2, including the acquisition of phonetics, vocabulary, grammar, and pragmatics. Of these language skills, vocabulary acquisition is crucial for language learning. Several studies have posited that L2 learners mostly learn new words incidentally~\cite{ramos2015incidental, Nation_2001}. This suggests that an individual might gradually master a word or a set of words in an unconscious manner. This leads to a phenomenon:

\begin{phenomenon}
In Second Language Acquisition, human individuals typically expand their vocabulary gradually, in a fashion of incremental learning rather than an instantaneous acquisition.
\end{phenomenon}

A formal description of levels of language development is laid out in the Common European Framework of Reference for Languages (CEFR)~\footnote{The Common European Framework of Reference for Languages (CEFR) is a standard developed by the European Commission and officially published in 2001, with a revised edition in 2003. The framework serves as a guideline for language teaching and assessment across European Union countries, aiming to provide a common foundation and reference for curriculum design, syllabus development, language testing, and textbook compilation in Europe.}. Table~\ref{tab:cefr} (show in Appendix~\ref{appdeix:cefr}) showcases the required number of vocabulary size for different CEFR levels. The CEFR provides detailed descriptions of the skills language learners must achieve to effectively communicate. This can be taken as evidence of the progressive nature of vocabulary acquisition. 

\subsection{Problem Definition: Second Language Acquisition for LLMs}

\paragraph{Language adaptation}

The focus on developing large-scale open-source language models for high-resource languages like English and Chinese has unintentionally marginalized low-resource languages, despite there being about 7,000 languages in use globally. The lack of data and computational resources makes it challenging to develop effective models for these languages. A common practice is to enhance existing models by adding specialized data for these underrepresented languages~\cite{cui2023efficient,huang2024acegpt,zhao2024llama}, \textit{a.k.a}, language adaptation.

\paragraph{Vocabulary expansion in language adaptation}
As a preliminary study, we identified Arabic tokens from LLaMA2 vocabulary using regular expressions. It was observed that LLaMA2 vocabulary only includes the basic characters of the Arabic language, resulting in relatively slow encoding and decoding speeds compared to English. 
During domain adaptation, it is crucial for vocabulary expansion for the second language, since it could significantly speed up decoding speeds as the number of decoded tokens is reduced due to the adapted vocabulary.
Furthermore, although augmenting the existing vocabulary with tokens from additional languages, followed by training on corresponding language corpora, appears to be a logical strategy, empirical evidence suggests that the gains from this method are modest. 
This insight underscores the complexity of enhancing support for low-resource languages within the framework of current large-scale language models.

\paragraph{Research question} Therefore, inspired by the  humans'  Second Language Acquisition, we argue for  
\begin{quote}
 \textit{Is it beneficial to  adopt progressive vocabulary learning in language adaptation of LLMs?}    
\end{quote}

\section{Methodology: Progressive Vocabulary Expansion for Language Adaptation}
\label{sec:solution}
Conventional Byte Pair Encoding (BPE) algorithms first create a complete vocabulary by iteratively merging the most frequent character pairs from a corpus, and then commence model training with this static vocabulary. This approach, while effective for monolingual models, presents challenges when adapting to new languages as it offers no mechanism for vocabulary evolution during training.
To address this limitation, we propose Progressive Vocabulary Expansion.

\subsection{Incremental Byte Pair Encoding Algorithm}
In contrast to standard BPE algorithms~\cite{sennrich2015neural} that use a static vocabulary, we introduce \textbf{Incremental Byte Pair Encoding~(I-BPE)} that dynamically augments the vocabulary during training. This approach mirrors human language acquisition, where vocabulary growth occurs simultaneously with deepening language comprehension. Algorithm~\ref{algo:1} outlines our method.

\begin{algorithm}[htb]
\caption{Incremental Byte Pair Encoding (I-BPE) Algorithm}
\label{algo:1}
\begin{algorithmic}[1]
    \State \textbf{Input:} (1) Initial vocabulary $V$; (2) Vocabulary size at each stage: $s_0, s_1, \ldots, s_n$; (3) Proportion of training corpus for newly added tokens at each stage: $r_0, r_1, \ldots, r_n$;
    
    \State \textbf{Output:} Final vocabulary $V$ for model training and application
    
    \For{$i = 0$ to $n$}
        \While{$|V| < s_i$}
            \State Compute frequency of all adjacent token pairs in $V$
            \State Identify the most frequent token pair $P_{freq}$
            \State Merge $P_{freq}$ into a new token $T_{new}$
            \State Add $T_{new}$ to vocabulary $V$
        \EndWhile
        \State Adjust corpus proportion for newly added tokens to $r_i$
        \State Train model with the updated vocabulary $V$ until convergence
    \EndFor
    
    \State \textbf{Return} Finalized vocabulary $V$
\end{algorithmic}
\end{algorithm}

The key innovation of I-BPE is its staged approach to vocabulary expansion. At each stage, we expand the vocabulary to a predetermined size, then train the model while gradually increasing the proportion of data corresponding to newly added tokens. This approach significantly reduces out-of-vocabulary (OOV) tokens at each training phase, enabling the model to incorporate new linguistic elements while preserving previously acquired knowledge.

\subsection{Expansion Strategies Comparison}
\begin{figure}[h]
\vspace{-10pt}
\centering
\includegraphics[width=0.5\textwidth]{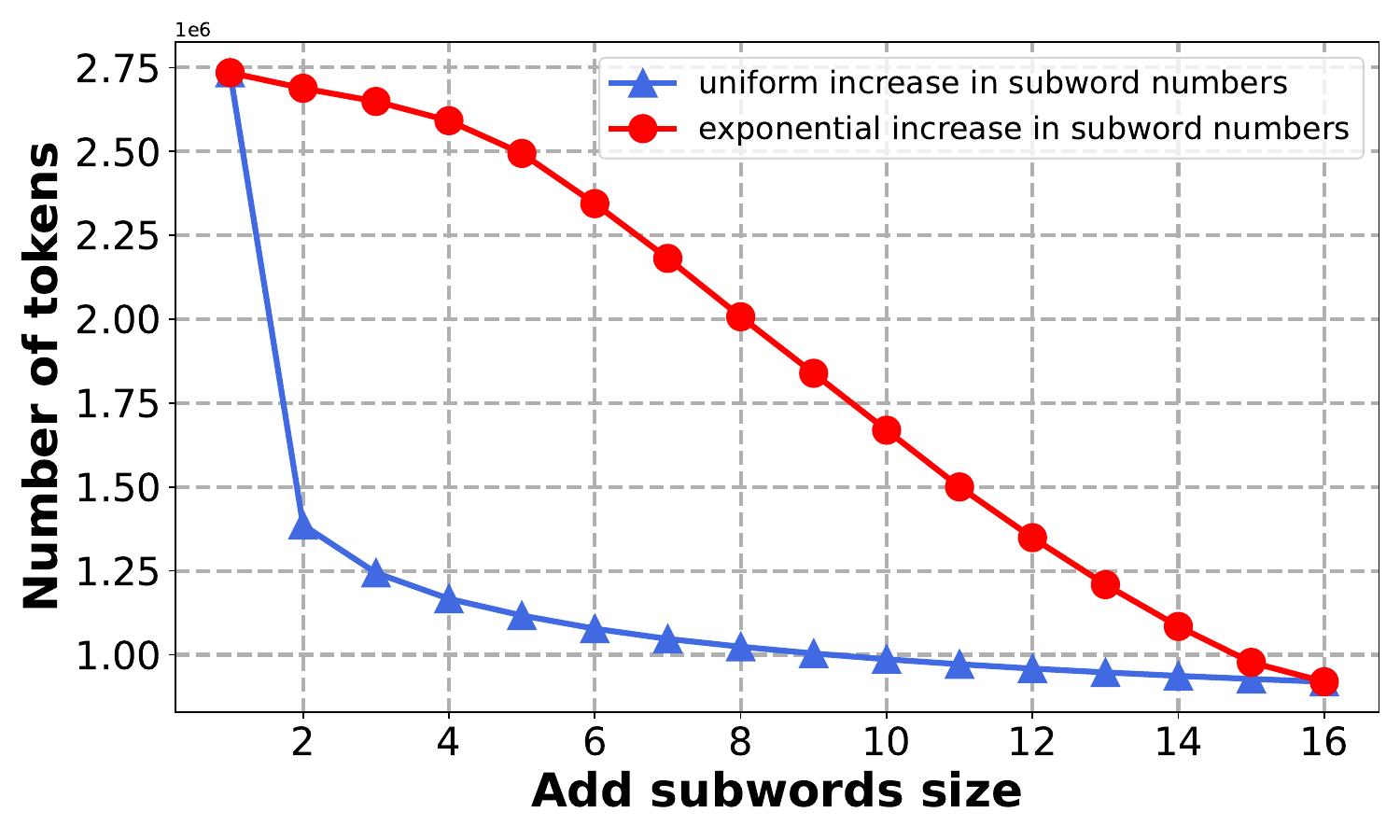}
\vspace{-10pt}
\caption{
Compression ratio comparison between uniform and exponential vocabulary expansion strategies.
}
\label{fig:vocabspeedup}
\end{figure}

For implementing vocabulary expansion, we investigated two principled strategies (illustrated in Figure~\ref{fig:vocabspeedup}):

\textbf{Uniform Expansion}: Adds a fixed number $K$ of tokens at each stage, resulting in $(T-1) \times K$ total additions over $T$ stages.
    
 \textbf{Exponential Expansion}: Doubles the number of new tokens at each stage following the sequence $\{0, 1, 2, \cdots, 2^{T-2}\}$, mimicking human language acquisition patterns.

Our comparative analysis using an identical Arabic corpus through 16 progressive stages revealed crucial differences between the two approaches. As shown in Figure~\ref{fig:vocabspeedup} and detailed in Table~\ref{tab:oov rate comparison}, uniform expansion causes abrupt improvements in compression ratio during early stages followed by diminishing returns. This sudden introduction of many tokens creates training instabilities and risks catastrophic forgetting as the model's representation space must rapidly accommodate numerous new tokens simultaneously.

Exponential expansion, however, offers critical advantages through its graduated approach: it provides superior training stability as the gradual introduction of tokens allows smooth adaptation of the model's representation space; it maintains consistent OOV ratios throughout training, preventing sudden vocabulary distribution shifts; and it achieves significant computational efficiency with a 3 times reduction in sequence length compared to the original LLaMA tokenizer. Based on these findings, we implemented exponential expansion with 12,800 Arabic subwords across 16 stages ($\log_2^{12800}$), representing the optimal saturation point for compression ratio improvement.

\subsection{Compression Ratios and Tokenizer Evaluation}
\jianqing{To rigorously assess the effectiveness of our vocabulary expansion approach, we conducted a comprehensive comparative evaluation of tokenization performance across multiple leading models. Using an identical Arabic corpus of 39 million words, we analyzed how different tokenizers processed Arabic text, with particular attention to efficiency metrics that impact both performance and computational requirements.

\begin{table*}[h]
\centering
\resizebox{\textwidth}{!}{
\begin{tabular}{lccccc}
\toprule
\textbf{Tokenizer} & \textbf{Total Words} & \textbf{Total Tokens} & \textbf{Subword Fertility} & \textbf{Ratio of Words Unbroken} & \textbf{Rényi Efficiency} \\
\midrule
LLaMA2(AceGPT) & 39,006,442 & 210,027,671 & 5.3844 & 0.0183 & 0.7731 \\
Bloomz & 39,006,442 & 80,617,499 & 2.0668 & 0.3176 & 0.7709 \\
Mistral & 39,006,442 & 206,082,344 & 5.2833 & 0.0185 & 0.7928 \\
Jais & 39,006,442 & 75,126,494 & 1.9260 & 0.3895 & 0.7343 \\
Our model& 39,006,442 & 66,554,771 & 1.7063 & 0.6323 & 0.7491 \\
\bottomrule
\end{tabular}}
\caption{Comprehensive tokenizer evaluation using standard metrics across different models.}
\label{tab:tokenizer-eval}
\end{table*}

The results in Table~\ref{tab:tokenizer-eval} reveal notable differences in how these models handle Arabic text. Our model achieved a token compression ratio of 0.3174 (ratio of tokens to original text size), representing a 68\% improvement over LLaMA2's baseline, which directly enhances inference speed and reduces memory requirements. We evaluated several key metrics established in recent tokenizer evaluation literature:

\begin{enumerate}
\item \textbf{Subword Fertility \cite{rust2021good,moosa2023transliteration}:} This metric measures the average number of tokens per word. Our model achieves the most optimal fertility (1.7063), approximately 3 times more efficient than LLaMA2 (5.3844) and Mistral (5.2833), while also outperforming multilingual models like Bloomz \cite{muennighoff2022crosslingual} (2.0668) and Jais (1.9260) that were specifically designed with Arabic support.

\item \textbf{Word Integrity \cite{moosa2023transliteration}:} For Arabic's rich morphology, preserving words as single tokens is vital. Our model achieves 63.23\% word integrity, far exceeding LLaMA2 (1.8\%) and outperforming Arabic-optimized models like Jais (38.95\%) and Bloomz (31.76\%).

\item \textbf{Total Tokens:} For the identical test corpus, our model requires only 66.55 million tokens, compared to LLaMA2's 210.03 million,a reduction of approximately 68\% that translates directly to memory savings and computational efficiency in both training and inference phases.

\item \textbf{Rényi Efficiency \cite{zouhar2023tokenization}:} This information-theoretic measure (higher values indicate better vocabulary utilization) shows our model (0.7491) achieves comparable efficiency to LLaMA2 (0.7731) despite its much lower token count, indicating efficient use of vocabulary space while maintaining high word integrity.
\end{enumerate}

The comparative analysis indicates that our model achieves an optimal equilibrium between morphological preservation and computational efficiency. Although models such as LLaMA2 and Mistral exhibit marginally superior Rényi Efficiency coefficients, this advantage is offset by substantial deficiencies in word integrity preservation and significantly elevated token densities. When compared with models specifically optimized for Arabic processing, such as Jais and Bloomz, our model consistently demonstrates superior performance across the majority of evaluation metrics, validating the efficacy of the progressive vocabulary expansion methodology for non-Latin script languages.
}

\section{Training Methodology}
Based on the Progressive Vocabulary Expansion methodology described above, we develop AraLLaMA, an Arabic Large Language Model that implements our proposed I-BPE algorithm. In this section, we detail the AraLLaMA training process, including data engineering and training specifics.

\subsection{Data Engineering}
\label{sec:data}

\paragraph{Pre-training Corpora}
Our pre-training dataset comprises both Arabic and English corpora. We employ an array of Arabic corpora encompassing multiple categories as delineated in Table \ref{tab:pretrain data} (shown in Appendix~\ref{appdeix:data distribution}). These include filtered versions of Common Crawl, WebText, and Wikipedia1 sourced from Joud and BAAI, all of which were subjected to additional cleaning processes. Moreover, we gather and purify additional corpora, namely Wikipedia2, Books, and Newspapers. The English corpus is sourced from SlimPajama~\cite{cerebras2023slimpajama} and Proof-Pile-2~\cite{azerbayev2023llemma}.

\paragraph{Data for Instruction Tuning}
After pre-training, we aim to elicit the knowledge out of \model via instruction tuning. Inspired by GLAN~\cite{glan}, we introduce ALAN (Arabic Instruction Tuning for Language Models). This method utilizes specific topics targeting Arabic knowledge to generate a vast amount of synthetic instruction data.

Specifically, we identified 127 critical topics within Arabic culture, science, and engineering as our focus. ALAN decomposes these topics into a structured hierarchy of fields, sub-fields, and individual disciplines. For each discipline, ALAN compiles a comprehensive list of subjects and designs a syllabus with specific knowledge points for each one. Using {\tt GPT-4-0613}, ALAN has generated 11,430 subjects and 244,812 detailed knowledge points. We provide more concrete examples in Appendix \ref{app:alan_examples}.

Armed with this extensive collection of subjects and knowledge points, we direct the LLM to create questions and answers related to these knowledge concepts. The syllabus consists of several lectures, each with 2 to 5 knowledge points. To diversify the knowledge base, we combine knowledge points from both the same and different lectures to produce diverse instructions and answers. Additionally, to vary the instruction types, the LLM generates three kinds of questions at random: multiple-choice, open-ended, and coding questions. In total, we've generated 733,419 instruction tuning data pieces. 

We also incorporated instruction tuning data from previous AceGPT projects. These include Quora-Arabic, Alpaca-Arabic \cite{alpaca}, Code-Alpaca-Arabic \cite{codealpaca}, Evol-Instruct-Arabic~\cite{xu2023wizardlm}, and ShareGPT data.

\subsection{Training details}
\label{sec:training}
The refined methodology for LLaMA2 model's vocabulary expansion incorporated 12,800 Arabic subwords derived through the I-BPE algorithm. The initialization procedure for each new token employed decomposition into constituent subwords from the original LLaMA2 vocabulary, with embedding initialization achieved via averaging the embeddings of these component tokens. This initialization strategy preserves semantic relationships between new and existing tokens, thereby enhancing training stability and facilitating vocabulary integration.

The training procedure was structured into 16 distinct stages~\footnote{Although incremental token addition is theoretically feasible, a staged implementation ($N=16$) was chosen to simplify data preparation.}, each processing 30B tokens, culminating in a total corpus of 480B tokens. A cosine annealing schedule governed the proportion of Arabic to English content, with Arabic representation increasing systematically from 30\% to 90\% across stages. This progressive exposure enables gradual adaptation to Arabic linguistic structures while preserving cross-lingual transfer capabilities via continued English exposure. Mathematical and programming content was maintained at a consistent 5\% throughout all stages to ensure robust inference capabilities in these domains (see Appendix~\ref{app:mixture}). The final training distribution comprised approximately 251.4B Arabic tokens and 204.6B English tokens.

The pre-training framework consisted of two principal epochs: an initial epoch employing vocabulary annealing for data distribution optimization, followed by a secondary epoch utilizing the fully refined vocabulary. An additional 20B tokens of training data were processed subsequent to vocabulary expansion to further enhance model performance. Each training phase implemented a discrete cosine learning rate schedule with warm-up period, producing a vocabulary-specific model at its conclusion, thereby rendering each phase functionally independent.

This stage-wise approach facilitates systematic integration of new tokens, enabling the model to adapt to evolving data representations while developing comprehensive understanding of linguistic patterns. The graduated modulation of language distribution—progressively increasing Arabic content while decreasing English representation—optimizes the model's capacity to process Arabic while maintaining cross-lingual capabilities.

The implementation utilized LLaMA2 architecture in 7B and 13B parameter configurations, trained on 2,368 Ascend 910A processors. Training durations were 7 and 11 days for the 7B and 13B models, respectively. The computational configuration employed model parallelism of degree 2 and pipeline parallelism of degree 4. Optimization was conducted using AdamW with 4,096-token context length. Each training stage utilized a cosine learning rate scheduler initialized at 1e-5 and decaying to 2e-6, with a 15\% warm-up interval. Gradient accumulation factor 8 yielded an effective batch size of 4,736, enabling processing of approximately 0.019B tokens per batch.

\section{Experiments}

\subsection{Experimental settings}

\paragraph{Benchmarking Datasets}
To assess world knowledge, we employ four widely used benchmarks. \textbf{\textit{MMLU}} (Measuring Massive Multitask Language Understanding)~\cite{hendrycks2021measuring} evaluates knowledge acquired during pretraining across a broad range of subjects; we utilize both the original English version and the Arabic version introduced in~\cite{huang2024acegpt} to ensure multilingual coverage. \textbf{\textit{RACE}} (Reading Comprehension from Examinations) serves as a large-scale English reading comprehension benchmark that focuses on educational knowledge. \textbf{\textit{EXAMS}} (Multi-subject High School Examinations Dataset for Cross-lingual and Multilingual Question Answering) further expands coverage by including subject-diverse exam questions drawn from multiple languages and curricula. \textbf{\textit{ArabicMMLU}} complements these by providing an Arabic-specific variant of MMLU, tailored to reflect regional knowledge across various Arab countries and subjects. Beyond general knowledge evaluation, we also examine cultural and value alignment using \textbf{\textit{ACVA-all}} and \textbf{\textit{ACVA-clean}}, which focus on Arabic cultural relevance and localization. To comprehensively evaluate inference and reasoning abilities, we translate two commonsense reasoning benchmarks of varying difficulty—\textbf{\textit{BoolQ}} and \textbf{\textit{ARC-Challenge (ARC-C)}}—into Arabic, allowing for consistent cross-lingual assessment.

To ensure a fair comparison of candidate models, we adhere to the settings established for each benchmark separately. Furthermore, for translated benchmarks, we utilize the generation approach evaluation method as outlined in~\cite{huang2024acegpt}. Specifically, we employed {\tt GPT-3.5-Turbo-1106} to translate datasets from English to Arabic for benchmarks that were not originally in Arabic.

\paragraph{Baselines} 
To compare LLMs trained or available in Arabic, we have selected several prominent Arabic LLMs or multilingual LLMs as baselines for comparison:
\textbf{(1)~AceGPT-[7B,13B]}: This set includes fully fine-tuned generative text models based on LLaMA2, specifically customized for the Arabic domain.
\textbf{(2)~Mistral-7B-Instruct-v0.2~\cite{jiang2023mistral}}: The fine-tuned model achieves a balance between performance and efficiency.
\textbf{(3)~Jais-[13B,30B]~\cite{sengupta2023jais}}: A pre-trained bilingual large language model designed for both Arabic and English.
\textbf{(4)~Bloom-[7B]}: A multilingual language model extensively trained on diverse textual data, allowing it to produce fluent text in 46 languages and 13 programming languages.
\textbf{(5)~LLaMA2-[7B,13B]}: A popular and competitive baseline model in the general domain.
\textbf{(6)~OpenAI GPT}: This includes GPT4 and ChatGPT, closed-source LLMs also strong at multilingual tasks.

\subsection{Evaluation Results}
\paragraph{Evaluation on Base Models} In our study, the performance of base models was assessed on two Arabic-specific MMLU datasets: Arabic MMLU translate \cite{huang2024acegpt} and ArabicMMLU \cite{koto2024arabicmmlu}. The left side of Table \ref{tab:mmlu-result} details the models' accuracies on the Arabic MMLU translate dataset within a few-shot setting. It is evident from the data that the \modelSmallBase and \modelLargeBase models exhibit superior accuracy rates compared to models of similar scale. Notably, the \modelLargeBase model outperforms the Jais-30B model, which has a significantly larger parameter count.

Additionally, the right side of Table \ref{tab:mmlu-result} presents the accuracy results of models in a zero-shot learning scenario. Here again, the \model models stand out for their exceptional performance, even when compared to models with similar parameter sizes. In particular, the \modelLargeBase model demonstrates a marked advantage over the Jais-30B-base model, notwithstanding the latter's larger size in terms of parameters.

These findings affirm the effectiveness of the \model models, developed through an annealing algorithm to expand the vocabulary, highlighting our methodology as a productive strategy for enhancing large models' adaptability to less prevalent languages. This contribution significantly advances the field of language model adaptation, offering a novel avenue for enriching language technology's inclusivity and depth.

\begin{table*}[htb]
\setlength{\tabcolsep}{2pt}
\centering
\footnotesize
\resizebox{0.9\textwidth}{!}{
\begin{tabular}{l|cccc|c|ccccc|c|c}
\toprule
\textbf{Models}                & \multicolumn{5}{c}{Arabic-trans MMLU~\cite{huang2024acegpt}}& \multicolumn{6}{c}{ArabicMMLU~\cite{koto2024arabicmmlu}} &\textbf{Total}  \\
           &  STEM & \makecell[c]{Human-\\ities} &\makecell[c]{Social \\ Sciences}   & Others &\textbf{Avg.}& STEM & \makecell[c]{Social \\ Sciences} &  \makecell[c]{Human-\\ities}   &  \makecell[c]{Arabic \\ Language}  & Other  & \textbf{Avg.} & \textbf{Avg.}\\
\midrule
Bloomz-7B-base    & \textbf{33.35} & 29.29 & 37.58 & 34.53 & 33.69&  - &-& -& -& -& -&-\\
LLaMA2-7B-base   & 30.30 & 29.33 & 27.46 & 30.78 & 29.47 &33.7 &32.8 &33.5& 28.4 &36.7 &33.4&31.43\\
AceGPT-7B-base   & 29.73 & 30.95 & 33.45 & 34.42 & 32.14& 35.4 &35.9& \textbf{36.2} &\textbf{31.1}& \textbf{41.7}& 36.3&34.22\\
\textbf{\modelSmallBase}  & 33.03 & \textbf{32.08} & \textbf{35.39} &\textbf{35.59} & \textbf{34.03}& \textbf{36.7}	&	\textbf{36.5}&	34.1 &30.0&	41.2 &\textbf{37.0} &\textbf{35.52}\\
\hline
LLaMA2-13B-base  & 32.94 & 32.30 & 33.42 & 37.27 & 33.76& 32.9 &35.0& 37.8 &35.8 &39.3 &36.1&34.93 \\
Jais-13B-base    & 30.51 & 31.25 & 33.74 & 33.43 & 33.76&  30.3 &31.4& 33.6 &28.1& 36.3 &32.2 &32.98\\
AceGPT-13B-base  & \textbf{36.60} & 38.74 & 43.76 & \textbf{42.72} & 40.45&  \textbf{42.7} &45.5 &48.3 &42.4 &50.7& 46.1 &43.28\\
\textbf{\modelLargeBase} & 36.13 &\textbf{40.07}  & \textbf{45.43} & 42.17& \textbf{40.95} & 42.4& \textbf{45.7}&\textbf{48.4}& \textbf{46.3}&\textbf{52.5}&\textbf{47.6}&\textbf{44.28}\\
\hline
Jais-30B-v1-base   & 32.67 & 30.67 & 42.13 & 39.60 & 36.27& 39.5 &45.6 &50.5& 34.6 &49.1 &44.8 &40.54\\
GPT-3.5 Turbo  & 43.38 & 44.12 & 55.57 & 53.21 & 49.07& 53.8& 57.0 &57.5& 57.6& 63.8 &57.7&53.39\\
\bottomrule
\end{tabular}}
\caption{
Evaluation of base models. We adopt a few-shot setting on Arabic-translated MMLU~\cite{huang2024acegpt} and a zero-shot setting with option logit probability in ArabicMMLU~\cite{koto2024arabicmmlu}. Numbers with the best performance are in \textbf{bold} in 7B and 13B groups.}
\label{tab:mmlu-result}
\end{table*}

\begin{table*}[htb]
\setlength{\tabcolsep}{2pt}
    \centering
    \footnotesize
    \resizebox{0.9\textwidth}{!}{
    \begin{tabular}{l|ccccccc|c|cc|c|c}
    \toprule
    \textbf{Models} & \multicolumn{8}{c|}{\textbf{Arabic}} & \multicolumn{3}{c|}{\textbf{English}} & \textbf{Total} \\
         & \makecell[c]{MMLU \\ (trans)}&\makecell[c]{MMLU \\ {\tiny \cite{koto2024arabicmmlu}}} & EXAMS & \makecell[c]{ACVA\\clean} & \makecell[c]{ACVA\\all} & \makecell[c]{BoolQ \\ (trans)} & \makecell[c]{ARC-C \\ (trans)} & \textbf{Avg.} & BoolQ & RACE & \textbf{Avg.} & \textbf{Avg.}  \\
        \midrule
        LLaMA2-7B-chat & 13.78 &33.40& 13.05 & 20.99 & 21.80 & 34.92 & 23.72 & 21.09 & 71.31 & 50.49 & 60.90 & 31.49 \\
        Phoenix-7b & 29.72&44.74 & 31.93 & 43.80 & 41.86 & 66.70 & 33.53 & 41.75 & 62.23 & 60.97 & 61.60 & 46.16 \\
        AceGPT-7B-chat & 30.69 &36.31& 33.73 & 53.87 & 53.07 & 60.70 & 38.05 & 43.77 & 54.74 & 53.97 & 54.36 & 46.12 \\
        Mistral-7B-Instruct-v0.2 & 27.93 & 41.44 & 21.56 & 64.56 & 63.47 & 60.18 & 35.67 & 44.97 & \textbf{84.53} & \textbf{73.17} & \textbf{78.85} & 52.50 \\
        \textbf{\modelSmallChat} & \textbf{45.77} &\textbf{56.62}& \textbf{43.69} & \textbf{69.46} & \textbf{70.86} & \textbf{72.45} & \textbf{60.49} & \textbf{59.90} & 75.78 & 72.13 & 73.96 & \textbf{63.02} \\
        \hline
        Jais-13B-chat & 19.52 &54.83& 19.71 & 66.75 & 61.41 & 41.25 & 11.95 & 39.34 & 28.13 & 20.08 & 24.10 & 35.96 \\
        LLaMA2-13B-chat & 8.92&36.12 & 16.11 &35.12 & 35.71 &  54.13 & 27.47 & 30.51 & 62.87 & 48.28 & 55.58 & 36.08 \\
        AceGPT-13B-chat & 35.59 &52.61& 38.72 & 70.82 & 70.21 & 66.85 & 44.20 & 54.14 & 60.55 & 45.22 & 52.88 & 53.86 \\
        \textbf{\modelLargeChat} & \textbf{47.33} &\textbf{61.70}& \textbf{48.37} & \textbf{76.90} & \textbf{76.37} & \textbf{69.33} & \textbf{63.99} & \textbf{63.42} & \textbf{83.67} & \textbf{80.82} & \textbf{82.24} & \textbf{67.61} \\
        \hline
        Jais-30B-chat-v1 & 38.12&59.33 & 40.45 & 74.46 & 72.41 & 73.76 & 50.94 & 58.49 & 65.05 & 75.26 & 70.16 & 61.09  \\
        Jais-30B-chat-v3 & 35.68&62.36 & 32.24 & 73.63 & 73.66 & {76.30} & 51.02 & 57.84 & 79.54 & {85.23} & {82.43} & 63.29  \\
        GPT-3.5 Turbo & {46.07}&57.72 & {45.63} & {74.45} & {76.88} & {76.12} & 60.24 & {62.44} & {85.32} & {84.65} & {84.99} & {67.45} \\
    \bottomrule
    \end{tabular}
    }
    \caption{Chat Models Evaluation in zero-shot setting. Numbers with best performance are in \textbf{bold} in 7B and 13B groups.}
    \label{tab:chat model evaluation}
\end{table*}

\paragraph{Evaluation on Chat Models} Table~\ref{tab:chat model evaluation} presents the comprehensive evaluation results across various benchmarks for the candidate models, spanning from Arabic to English. Overall, \model outperforms all baseline models in the Arabic tasks. Particularly noteworthy is its proficiency in knowledge-related evaluations such as Arabic-translated MMLU and EXAMS, surpassing other models by at least 1.3\%. This highlights the model's expertise in addressing Arabic knowledge-related questions. Additionally, \model demonstrates strong performance in tasks related to Arabic culture and value alignment.
In terms of commonsense reasoning, \model exhibits notable skills in tasks such as the translated versions of BoolQ and ARC-Challenge, showcasing its reasoning capabilities in Arabic. Beyond Arabic benchmarks, we also investigated the English proficiency of the models to determine whether specialization in one language affects performance in the other. The results indicate that the model maintains its English proficiency and displays robustness in multilingual assessments. It is noteworthy that the lower accuracy of the Jais is attributed to its refusal to answer for unknown reasons.

In a comprehensive evaluation of the ACVA dataset aimed at gauging the understanding of Arabic cultural nuances under a zero-shot setting, our \model models showcased unparalleled performance. The \modelLargeChat, in particular, stood out with exceptional Average F1 scores of 76.37\% and 76.90\% in ``all set" and "clean Set" categories, respectively, even outperforming the renowned GPT-3.5 Turbo in the "All set" category. This performance not only highlights the \model models' superior grasp of Arabic culture but also establishes them as leading figures among open-source models in this nuanced domain. Compared to other top-tier open-source contenders, including the Jais-30B-chat variants, the \modelLargeChat model's superior results. The instruction-following tests can be found in Appendix \ref{sec:instrucion-following}.

\section{More Analysis}

\begin{table*}[h]
\centering
\footnotesize
\resizebox{0.8\textwidth}{!}{
    \begin{tabular}{l|ccccc|c}
    \toprule
    Model         & STEM & \makecell{Social\\Sciences} & Humanities & \makecell{Arabic\\Language} & Other & \textbf{Avg.}  \\
    \midrule
    TinyLLaMA chat & 35.1 & 36.9            & 38.5       & 28.6            & 39.8  & 36.5 \\
    TinyLLaMA (VE) chat    & 35.3 & 39.7            & 40.1       & \textbf{33.8}            & 41.6  & 38.5 \\
    TinyLLaMA (PVE) chat     & \textbf{36.3} & \textbf{40.7}            & \textbf{44.2}       & 33.5            & \textbf{45.7}  & \textbf{40.7} \\
    \bottomrule
    \end{tabular}
}
\caption{Performance comparison on ArabicMMLU~\cite{koto2024arabicmmlu} across different domains.}
\label{tab:tinyllama-mmlu}
\end{table*}

\subsection{Ablation Study on Progressive Vocabulary Expansion}
\label{sec:ablation}

\juhao{To further demonstrate the effectiveness of progressive vocabulary expansion in downstream task adaptation, we conduct continuous pre-training on a 1B-parameter TinyLLaMA model~\cite{zhang2024tinyllama}, followed by supervised fine-tuning. More details on the experimental setup can be found in Appendix~\ref{appendix: details of ablation}.}

A comprehensive analysis is conducted by applying the same Supervised Fine-Tuning (SFT) protocol across three pre-training configurations: the baseline TinyLLaMA model, TinyLLaMA with Progressive Vocabulary Expansion (PVE), and TinyLLaMA with Vocabulary Expansion all at once (VE). The performance of these models is evaluated on the Arabic MMLU (see Table \ref{tab:tinyllama-mmlu}) and Arabic Vicuna-80 (see Table \ref{tab:tinyllama-vicuna80}) benchmarks. Experiment results demonstrate that vocabulary expansion significantly enhances model performance, with the PVE approach yielding superior results across various categories in the Arabic MMLU benchmark, achieving an average score of 40.7 compared to 38.5 for VE and 36.5 for the baseline model. Similarly, in the Arabic Vicuna-80 comparison, the PVE method led to the highest accuracy of 29.18\%, outperforming VE (22.61\%) and the baseline model (21.3\%). These results underscore the effectiveness of progressive vocabulary expansion in enhancing language model performance, particularly in complex language tasks.

\begin{table}[h]
\centering
\footnotesize
\begin{tabular}{lc}
\toprule
\textbf{Model} & \textbf{Accuracy (\%)} \\
\midrule
TinyLLaMA chat & 21.30 (\textit{baseline}) \\
TinyLLaMA (VE) chat & 22.61 (\textcolor{black!30!green}{$+ 1.31$}) \\
TinyLLaMA (PVE) chat & \textbf{29.18} (\textcolor{black!30!green}{$+ 7.88$}) \\
\bottomrule
\end{tabular}
\caption{Performance Comparison on Arabic Vicuna-80 Benchmark}
\label{tab:tinyllama-vicuna80}
\end{table}

\subsection{Benchmarking in English dataset}
We evaluated the accuracy of both base and chat models on the English MMLU dataset. As illustrated in Table \ref{tab:mmlu-result} (shown in Appendix \ref{enmmlu}), in the base model category, \modelPrefix's accuracy is slightly lower than that of the original LLaMA model but notably higher than the AceGPT model, which is also trained on the LLaMA architecture. This indicates that expanding Arabic capabilities via an annealing algorithm does not compromise the model's inherent English proficiency. This offers a viable solution for language transfer in large models. After undergoing SFT, \model achieves the highest accuracy among models of similar size and surpasses the Jais-30B model, which has a greater number of parameters.

\subsection{Decoding Efficiency Analysis}
We conducted a systematic evaluation of generation efficiency between LLaMA2 and AraLLaMA 7B chat models on Arabic text generation tasks. Each model was tested on standardized Arabic prompts with a maximum output length of 100 tokens. To ensure statistical reliability, we performed five independent trials and analyzed only Arabic language outputs, excluding any non-Arabic tokens from the performance calculations.

\begin{table}[h]
\centering
\footnotesize
\begin{tabular}{lcc}
\toprule
\textbf{Model} & \textbf{Tokens/Second} & \textbf{Words/Second} \\
\midrule
LLaMA2 & $29.68\pm0.04$ & $4.55\pm0.50$ \\
AraLLaMA & $30.12\pm0.06$ & $20.37\pm0.04$ \\
\bottomrule
\end{tabular}
\caption{Comparative analysis of generation speed between LLaMA2 and AraLLaMA on Arabic text.}
\label{tab:decoding-speed}
\end{table}

Table \ref{tab:decoding-speed} shows that while both models achieve similar token processing speeds (~30 tokens/second, $p > 0.05$), AraLLaMA generates words 4.5× faster. This efficiency gain (from $4.55\pm0.50$ to $20.37\pm0.04$ words/second) demonstrates the effectiveness of our vocabulary expansion approach. The improved word-level performance while maintaining similar token-level speeds indicates that our language-specific tokenization strategy successfully optimizes text generation for Arabic's morphological complexity.

\section{Conclusion}
Adapting large-scale models to less commonly spoken languages is fraught with challenges, notably the hurdles of knowledge transfer and the prevalence of OOV terms. We developed a novel annealing training algorithm to address these issues specifically for Arabic. This strategy methodically expands the vocabulary and employs a phased training process, leading to the development of the \model 7B and 13B models. Subsequent evaluations of both the base and chat configurations across diverse datasets have unequivocally established \modelPrefix's superior accuracy compared to peers within the same parameter range. Remarkably, the \model also exhibits robust performance advantages over models with significantly more parameters. The proven efficacy of our algorithm is supported by robust empirical evidence. Moving forward, we aim to further democratize access to advanced model technology by making our models, along with their code and datasets, openly available, thus making a meaningful contribution to the progress of the field.

\section*{Limitation}
This paper exhibits several limitations. Due to constraints in resources and budget, the models has not undergone evaluation by native Arabic speakers, which could affect its practicality and adoption. Consequently, its use is currently confined to academic research rather than online deployment. Additionally, the writing of this paper was supported by AI tools for grammar correction and refinement.

\section*{Acknowledgements}
This work was conducted under the platform of the KAUST-SRIBD Joint Lab on Scientific Computing and Machine Learning. We would like to acknowledge the support of Hetao Shenzhen-Hong Kong Science and Technology Innovation Cooperation Zone Project (No. HZQSWS-KCCYB-2024016), the Shenzhen Science and Technology Program (JCYJ20220818103001002), Shenzhen Doctoral Startup Funding (RCBS20221008093330065), Tianyuan Fund for Mathematics of National Natural Science Foundation of China (NSFC) (12326608), Shenzhen Key Laboratory of Cross-Modal Cognitive Computing (grant number ZDSYS20230626091302006), Shenzhen Stability Science Program 2023, Shenzhen Key Lab of Multi-Modal Cognitive Computing, and KAUST Baseline Research Fund.

\newpage

\bibliography{reference}

\begin{thebibliography}{45}
\providecommand{\natexlab}[1]{#1}

\bibitem[{Achiam et~al.(2023)Achiam, Adler, Agarwal, Ahmad, Akkaya, Aleman,
  Almeida, Altenschmidt, Altman, Anadkat et~al.}]{achiam2023gpt}
Josh Achiam, Steven Adler, Sandhini Agarwal, Lama Ahmad, Ilge Akkaya,
  Florencia~Leoni Aleman, Diogo Almeida, Janko Altenschmidt, Sam Altman,
  Shyamal Anadkat, et~al. 2023.
\newblock Gpt-4 technical report.
\newblock \emph{arXiv preprint arXiv:2303.08774}.

\bibitem[{Azerbayev et~al.(2023)Azerbayev, Schoelkopf, Paster, Santos, McAleer,
  Jiang, Deng, Biderman, and Welleck}]{azerbayev2023llemma}
Zhangir Azerbayev, Hailey Schoelkopf, Keiran Paster, Marco~Dos Santos, Stephen
  McAleer, Albert~Q. Jiang, Jia Deng, Stella Biderman, and Sean Welleck. 2023.
\newblock \href {https://arxiv.org/abs/2310.10631} {Llemma: An open language
  model for mathematics}.
\newblock \emph{Preprint}, arXiv:2310.10631.

\bibitem[{Bostrom and Durrett(2020)}]{bostrom2020byte}
Kaj Bostrom and Greg Durrett. 2020.
\newblock Byte pair encoding is suboptimal for language model pretraining.
\newblock \emph{arXiv preprint arXiv:2004.03720}.

\bibitem[{Chaudhary(2023)}]{codealpaca}
Sahil Chaudhary. 2023.
\newblock Code alpaca: An instruction-following llama model for code
  generation.
\newblock \url{https://github.com/sahil280114/codealpaca}.

\bibitem[{Chen et~al.(2023)Chen, Jiang, Chen, Wang, Yu, Chen, Zhang, Liang,
  Zhang, Zhang et~al.}]{chen2023phoenix}
Zhihong Chen, Feng Jiang, Junying Chen, Tiannan Wang, Fei Yu, Guiming Chen,
  Hongbo Zhang, Juhao Liang, Chen Zhang, Zhiyi Zhang, et~al. 2023.
\newblock Phoenix: Democratizing chatgpt across languages.
\newblock \emph{arXiv preprint arXiv:2304.10453}.

\bibitem[{Chiang et~al.(2023)Chiang, Li, Lin, Sheng, Wu, Zhang, Zheng, Zhuang,
  Zhuang, Gonzalez, Stoica, and Xing}]{vicuna2023}
Wei-Lin Chiang, Zhuohan Li, Zi~Lin, Ying Sheng, Zhanghao Wu, Hao Zhang, Lianmin
  Zheng, Siyuan Zhuang, Yonghao Zhuang, Joseph~E. Gonzalez, Ion Stoica, and
  Eric~P. Xing. 2023.
\newblock \href {https://lmsys.org/blog/2023-03-30-vicuna/} {Vicuna: An
  open-source chatbot impressing gpt-4 with 90\%* chatgpt quality}.

\bibitem[{Clark et~al.(2019)Clark, Lee, Chang, Kwiatkowski, Collins, and
  Toutanova}]{clark2019boolq}
Christopher Clark, Kenton Lee, Ming-Wei Chang, Tom Kwiatkowski, Michael
  Collins, and Kristina Toutanova. 2019.
\newblock Boolq: Exploring the surprising difficulty of natural yes/no
  questions.
\newblock In \emph{NAACL}.

\bibitem[{Clark et~al.(2018)Clark, Cowhey, Etzioni, Khot, Sabharwal, Schoenick,
  and Tafjord}]{allenai:arc}
Peter Clark, Isaac Cowhey, Oren Etzioni, Tushar Khot, Ashish Sabharwal, Carissa
  Schoenick, and Oyvind Tafjord. 2018.
\newblock Think you have solved question answering? try arc, the ai2 reasoning
  challenge.
\newblock \emph{arXiv:1803.05457v1}.

\bibitem[{Coady(1996)}]{coady1996l2}
James Coady. 1996.
\newblock L2 vocabulary acquisition through extensive reading.
\newblock In \emph{Second language vocabulary acquisition: A rationale for
  pedagogy}, pages 225--237. Cambridge University Press.

\bibitem[{Conover et~al.(2023)Conover, Hayes, Mathur, Xie, Wan, Shah, Ghodsi,
  Wendell, Zaharia, and Xin}]{DatabricksBlog2023DollyV2}
Mike Conover, Matt Hayes, Ankit Mathur, Jianwei Xie, Jun Wan, Sam Shah, Ali
  Ghodsi, Patrick Wendell, Matei Zaharia, and Reynold Xin. 2023.
\newblock \href
  {https://www.databricks.com/blog/2023/04/12/dolly-first-open-commercially-viable-instruction-tuned-llm}
  {Free dolly: Introducing the world's first truly open instruction-tuned llm}.

\bibitem[{Cui et~al.(2023)Cui, Yang, and Yao}]{cui2023efficient}
Yiming Cui, Ziqing Yang, and Xin Yao. 2023.
\newblock Efficient and effective text encoding for chinese llama and alpaca.
\newblock \emph{arXiv preprint arXiv:2304.08177}.

\bibitem[{Hardalov et~al.(2020)Hardalov, Mihaylov, Zlatkova, Dinkov, Koychev,
  and Nakov}]{EXAMS20}
Momchil Hardalov, Todor Mihaylov, Dimitrina Zlatkova, Yoan Dinkov, Ivan
  Koychev, and Preslav Nakov. 2020.
\newblock \href {https://doi.org/10.18653/v1/2020.emnlp-main.438} {{EXAMS:} {A}
  multi-subject high school examinations dataset for cross-lingual and
  multilingual question answering}.
\newblock In \emph{Proceedings of the 2020 Conference on Empirical Methods in
  Natural Language Processing, {EMNLP} 2020, Online, November 16-20, 2020},
  pages 5427--5444. Association for Computational Linguistics.

\bibitem[{Hendrycks et~al.(2021{\natexlab{a}})Hendrycks, Burns, Basart, Zou,
  Mazeika, Song, and Steinhardt}]{hendrycks2021measuring}
Dan Hendrycks, Collin Burns, Steven Basart, Andy Zou, Mantas Mazeika, Dawn
  Song, and Jacob Steinhardt. 2021{\natexlab{a}}.
\newblock \href {https://arxiv.org/abs/2009.03300} {Measuring massive multitask
  language understanding}.
\newblock \emph{Preprint}, arXiv:2009.03300.

\bibitem[{Hendrycks et~al.(2021{\natexlab{b}})Hendrycks, Burns, Basart, Zou,
  Mazeika, Song, and Steinhardt}]{MMLU21}
Dan Hendrycks, Collin Burns, Steven Basart, Andy Zou, Mantas Mazeika, Dawn
  Song, and Jacob Steinhardt. 2021{\natexlab{b}}.
\newblock \href {https://openreview.net/forum?id=d7KBjmI3GmQ} {Measuring
  massive multitask language understanding}.
\newblock In \emph{9th International Conference on Learning Representations,
  {ICLR} 2021, Virtual Event, Austria, May 3-7, 2021}. OpenReview.net.

\bibitem[{Huang et~al.(2024)Huang, Yu, Zhu, Sun, Cheng, Dingjie, Chen,
  Alharthi, An, He et~al.}]{huang2024acegpt}
Huang Huang, Fei Yu, Jianqing Zhu, Xuening Sun, Hao Cheng, Song Dingjie,
  Zhihong Chen, Mosen Alharthi, Bang An, Juncai He, et~al. 2024.
\newblock Acegpt, localizing large language models in arabic.
\newblock In \emph{Proceedings of the 2024 Conference of the North American
  Chapter of the Association for Computational Linguistics: Human Language
  Technologies (Volume 1: Long Papers)}, pages 8139--8163.

\bibitem[{Jiang et~al.(2023)Jiang, Sablayrolles, Mensch, Bamford, Chaplot,
  de~las Casas, Bressand, Lengyel, Lample, Saulnier, Lavaud, Lachaux, Stock,
  Scao, Lavril, Wang, Lacroix, and Sayed}]{jiang2023mistral}
Albert~Q. Jiang, Alexandre Sablayrolles, Arthur Mensch, Chris Bamford,
  Devendra~Singh Chaplot, Diego de~las Casas, Florian Bressand, Gianna Lengyel,
  Guillaume Lample, Lucile Saulnier, Lélio~Renard Lavaud, Marie-Anne Lachaux,
  Pierre Stock, Teven~Le Scao, Thibaut Lavril, Thomas Wang, Timothée Lacroix,
  and William~El Sayed. 2023.
\newblock \href {https://arxiv.org/abs/2310.06825} {Mistral 7b}.
\newblock \emph{Preprint}, arXiv:2310.06825.

\bibitem[{Koto et~al.(2024)Koto, Li, Shatnawi, Doughman, Sadallah, Alraeesi,
  Almubarak, Alyafeai, Sengupta, Shehata et~al.}]{koto2024arabicmmlu}
Fajri Koto, Haonan Li, Sara Shatnawi, Jad Doughman, Abdelrahman~Boda Sadallah,
  Aisha Alraeesi, Khalid Almubarak, Zaid Alyafeai, Neha Sengupta, Shady
  Shehata, et~al. 2024.
\newblock Arabicmmlu: Assessing massive multitask language understanding in
  arabic.
\newblock \emph{arXiv preprint arXiv:2402.12840}.

\bibitem[{Krashen(1981)}]{krashen1981second}
Stephen Krashen. 1981.
\newblock Second language acquisition.
\newblock \emph{Second Language Learning}, 3(7):19--39.

\bibitem[{Krashen(1982)}]{krashen1982principles}
Stephen Krashen. 1982.
\newblock \emph{Principles and practice in second language acquisition}.
\newblock Pergamon Press.

\bibitem[{Kudo(2018)}]{kudo2018subword}
Taku Kudo. 2018.
\newblock Subword regularization: Improving neural network translation models
  with multiple subword candidates.
\newblock \emph{arXiv preprint arXiv:1804.10959}.

\bibitem[{Lai et~al.(2017)Lai, Xie, Liu, Yang, and Hovy}]{lai-etal-2017-race}
Guokun Lai, Qizhe Xie, Hanxiao Liu, Yiming Yang, and Eduard Hovy. 2017.
\newblock \href {https://doi.org/10.18653/v1/D17-1082} {{RACE}: Large-scale
  {R}e{A}ding comprehension dataset from examinations}.
\newblock In \emph{Proceedings of the 2017 Conference on Empirical Methods in
  Natural Language Processing}, pages 785--794, Copenhagen, Denmark.
  Association for Computational Linguistics.

\bibitem[{Le{\'o}n~Romero et~al.(2016)}]{leon2016spaced}
H{\'e}ctor~Daniel Le{\'o}n~Romero et~al. 2016.
\newblock Spaced retrieval practice applied to vocabulary learning in secondary
  education.

\bibitem[{Li et~al.(2024)Li, Dong, Tang, Wang, Zhang, Huang, Huang, Huang,
  Huang, Zhang et~al.}]{glan}
Haoran Li, Qingxiu Dong, Zhengyang Tang, Chaojun Wang, Xingxing Zhang, Haoyang
  Huang, Shaohan Huang, Xiaolong Huang, Zeqiang Huang, Dongdong Zhang, et~al.
  2024.
\newblock Synthetic data (almost) from scratch: Generalized instruction tuning
  for language models.
\newblock \emph{arXiv preprint arXiv:2402.13064}.

\bibitem[{Moosa et~al.(2023)Moosa, Akhter, and
  Habib}]{moosa2023transliteration}
Ibraheem~Muhammad Moosa, Mahmud~Elahi Akhter, and Ashfia~Binte Habib. 2023.
\newblock Does transliteration help multilingual language modeling?
\newblock In \emph{Findings of the Association for Computational Linguistics:
  EACL 2023}, pages 670--685, Dubrovnik, Croatia. Association for Computational
  Linguistics.

\bibitem[{Muennighoff et~al.(2022)Muennighoff, Wang, Sutawika, Roberts,
  Biderman, Scao, Bari, Shen, Yong, Schoelkopf
  et~al.}]{muennighoff2022crosslingual}
Niklas Muennighoff, Thomas Wang, Lintang Sutawika, Adam Roberts, Stella
  Biderman, Teven~Le Scao, M~Saiful Bari, Sheng Shen, Zheng-Xin Yong, Hailey
  Schoelkopf, et~al. 2022.
\newblock Crosslingual generalization through multitask finetuning.
\newblock \emph{arXiv preprint arXiv:2211.01786}.

\bibitem[{Nakata(2015)}]{nakata2015effects}
Tatsuya Nakata. 2015.
\newblock Effects of expanding and equal spacing on second language vocabulary
  learning: Does gradually increasing spacing increase vocabulary learning?
\newblock \emph{Studies in Second Language Acquisition}, 37(4):677--711.

\bibitem[{Nation(2001)}]{Nation_2001}
I.~S.~P. Nation. 2001.
\newblock \emph{Learning Vocabulary in Another Language}.
\newblock Cambridge Applied Linguistics. Cambridge University Press.

\bibitem[{Nation and Nation(2001)}]{nation2001learning}
Ian~SP Nation and I.~S.~P. Nation. 2001.
\newblock \emph{Learning vocabulary in another language}, volume~10.
\newblock Cambridge University Press.

\bibitem[{Nguyen et~al.(2023)Nguyen, Zhang, Li, Aljunied, Tan, Cheng, Chen,
  Deng, Yang, Liu et~al.}]{nguyen2023seallms}
Xuan-Phi Nguyen, Wenxuan Zhang, Xin Li, Mahani Aljunied, Qingyu Tan, Liying
  Cheng, Guanzheng Chen, Yue Deng, Sen Yang, Chaoqun Liu, et~al. 2023.
\newblock Seallms--large language models for southeast asia.
\newblock \emph{arXiv preprint arXiv:2312.00738}.

\bibitem[{Ramos and Dario(2015)}]{ramos2015incidental}
Restrepo Ramos and Falcon Dario. 2015.
\newblock Incidental vocabulary learning in second language acquisition: A
  literature review.
\newblock \emph{Profile Issues in TeachersProfessional Development},
  17(1):157--166.

\bibitem[{Rust et~al.(2021)Rust, Pfeiffer, Vuli\'{c}, Ruder, and
  Gurevych}]{rust2021good}
Phillip Rust, Jonas Pfeiffer, Ivan Vuli\'{c}, Sebastian Ruder, and Iryna
  Gurevych. 2021.
\newblock How good is your tokenizer? on the monolingual performance of
  multilingual language models.
\newblock In \emph{Proceedings of the 59th Annual Meeting of the Association
  for Computational Linguistics and the 11th International Joint Conference on
  Natural Language Processing (Volume 1: Long Papers)}, pages 3118--3135.
  Association for Computational Linguistics.

\bibitem[{Salesky et~al.(2020)Salesky, Runge, Coda, Niehues, and
  Neubig}]{salesky2020optimizing}
Elizabeth Salesky, Andrew Runge, Alex Coda, Jan Niehues, and Graham Neubig.
  2020.
\newblock Optimizing segmentation granularity for neural machine translation.
\newblock \emph{Machine Translation}, 34(1):41--59.

\bibitem[{Sengupta et~al.(2023)Sengupta, Sahu, Jia, Katipomu, Li, Koto, Afzal,
  Kamboj, Pandit, Pal et~al.}]{sengupta2023jais}
Neha Sengupta, Sunil~Kumar Sahu, Bokang Jia, Satheesh Katipomu, Haonan Li,
  Fajri Koto, Osama~Mohammed Afzal, Samta Kamboj, Onkar Pandit, Rahul Pal,
  et~al. 2023.
\newblock Jais and jais-chat: Arabic-centric foundation and instruction-tuned
  open generative large language models.
\newblock \emph{arXiv preprint arXiv:2308.16149}.

\bibitem[{Sennrich et~al.(2015)Sennrich, Haddow, and
  Birch}]{sennrich2015neural}
Rico Sennrich, Barry Haddow, and Alexandra Birch. 2015.
\newblock Neural machine translation of rare words with subword units.
\newblock \emph{arXiv preprint arXiv:1508.07909}.

\bibitem[{Soboleva et~al.(2023)Soboleva, Faisal, R, Joel, and
  Nolan}]{cerebras2023slimpajama}
Daria Soboleva, Al-Khateeb Faisal, Myers Robert Steeves~Jacob R, Hestness Joel,
  and Dey Nolan. 2023.
\newblock \href {https://huggingface.co/datasets/cerebras/SlimPajama-627B}
  {{SlimPajama: A 627B token cleaned and deduplicated version of RedPajama}}.

\bibitem[{Taori et~al.(2023)Taori, Gulrajani, Zhang, Dubois, Li, Guestrin,
  Liang, and Hashimoto}]{alpaca}
Rohan Taori, Ishaan Gulrajani, Tianyi Zhang, Yann Dubois, Xuechen Li, Carlos
  Guestrin, Percy Liang, and Tatsunori~B. Hashimoto. 2023.
\newblock Stanford alpaca: An instruction-following llama model.
\newblock \url{https://github.com/tatsu-lab/stanford_alpaca}.

\bibitem[{Touvron et~al.(2023{\natexlab{a}})Touvron, Lavril, Izacard, Martinet,
  Lachaux, Lacroix, Rozi{\`e}re, Goyal, Hambro, Azhar
  et~al.}]{touvron2023llama1}
Hugo Touvron, Thibaut Lavril, Gautier Izacard, Xavier Martinet, Marie-Anne
  Lachaux, Timoth{\'e}e Lacroix, Baptiste Rozi{\`e}re, Naman Goyal, Eric
  Hambro, Faisal Azhar, et~al. 2023{\natexlab{a}}.
\newblock Llama: Open and efficient foundation language models.
\newblock \emph{arXiv preprint arXiv:2302.13971}.

\bibitem[{Touvron et~al.(2023{\natexlab{b}})Touvron, Martin, Stone, Albert,
  Almahairi, Babaei, Bashlykov, Batra, Bhargava, Bhosale
  et~al.}]{touvron2023llama}
Hugo Touvron, Louis Martin, Kevin Stone, Peter Albert, Amjad Almahairi, Yasmine
  Babaei, Nikolay Bashlykov, Soumya Batra, Prajjwal Bhargava, Shruti Bhosale,
  et~al. 2023{\natexlab{b}}.
\newblock Llama 2: Open foundation and fine-tuned chat models.
\newblock \emph{arXiv preprint arXiv:2307.09288}.

\bibitem[{{\"U}st{\"u}n et~al.(2024){\"U}st{\"u}n, Aryabumi, Yong, Ko, D'souza,
  Onilude, Bhandari, Singh, Ooi, Kayid et~al.}]{ustun2024aya}
Ahmet {\"U}st{\"u}n, Viraat Aryabumi, Zheng-Xin Yong, Wei-Yin Ko, Daniel
  D'souza, Gbemileke Onilude, Neel Bhandari, Shivalika Singh, Hui-Lee Ooi, Amr
  Kayid, et~al. 2024.
\newblock Aya model: An instruction finetuned open-access multilingual language
  model.
\newblock \emph{arXiv preprint arXiv:2402.07827}.

\bibitem[{Xu et~al.(2023)Xu, Sun, Zheng, Geng, Zhao, Feng, Tao, and
  Jiang}]{xu2023wizardlm}
Can Xu, Qingfeng Sun, Kai Zheng, Xiubo Geng, Pu~Zhao, Jiazhan Feng, Chongyang
  Tao, and Daxin Jiang. 2023.
\newblock Wizardlm: Empowering large language models to follow complex
  instructions.
\newblock \emph{arXiv preprint arXiv:2304.12244}.

\bibitem[{Xu et~al.(2020)Xu, Zhou, Gan, Zheng, and Li}]{xu2020vocabulary}
Jingjing Xu, Hao Zhou, Chun Gan, Zaixiang Zheng, and Lei Li. 2020.
\newblock Vocabulary learning via optimal transport for neural machine
  translation.
\newblock \emph{arXiv preprint arXiv:2012.15671}.

\bibitem[{Zhang et~al.(2024)Zhang, Zeng, Wang, and Lu}]{zhang2024tinyllama}
Peiyuan Zhang, Guangtao Zeng, Tianduo Wang, and Wei Lu. 2024.
\newblock Tinyllama: An open-source small language model.
\newblock \emph{arXiv preprint arXiv:2401.02385}.

\bibitem[{Zhang et~al.(2021)Zhang, Xu, and Zhang}]{zhang2021effects}
Songshan Zhang, Hai Xu, and Xian Zhang. 2021.
\newblock The effects of dictionary use on second language vocabulary
  acquisition: A meta-analysis.
\newblock \emph{International Journal of Lexicography}, 34(1):1--38.

\bibitem[{Zhao et~al.(2024)Zhao, Zhang, Zhang, Gui, and Huang}]{zhao2024llama}
Jun Zhao, Zhihao Zhang, Qi~Zhang, Tao Gui, and Xuanjing Huang. 2024.
\newblock Llama beyond english: An empirical study on language capability
  transfer.
\newblock \emph{arXiv preprint arXiv:2401.01055}.

\bibitem[{Zouhar et~al.(2023)Zouhar, Meister, Gastaldi, Du, Sachan, and
  Cotterell}]{zouhar2023tokenization}
Vil\'{e}m Zouhar, Clara Meister, Juan Gastaldi, Li~Du, Mrinmaya Sachan, and
  Ryan Cotterell. 2023.
\newblock Tokenization and the noiseless channel.
\newblock In \emph{Proceedings of the 61st Annual Meeting of the Association
  for Computational Linguistics (Volume 1: Long Papers)}, pages 5184--5207,
  Toronto, Canada. Association for Computational Linguistics.

\end{thebibliography}
\newpage
\appendix

\section{Related Work}
Our work primarily focuses on two key areas: low-resource language models and vocabulary expansion.

\paragraph{Low-resource language models}  
Recent efforts have centered on developing open-source LLMs as alternatives to proprietary models like GPT-3.5 Turbo and GPT-4 (\citealp{alpaca,vicuna2023,DatabricksBlog2023DollyV2,chen2023phoenix,sengupta2023jais}). These initiatives have expanded beyond English, addressing languages with fewer available resources and creating models specifically tailored to diverse linguistic landscapes (\citealp{chen2023phoenix,ustun2024aya}). SeaLLMs (\citealp{nguyen2023seallms}) are adapted from English-centric models by extending vocabulary and fine-tuning to better capture regional language complexities. Jais (\citealp{sengupta2023jais}) introduces a model trained from scratch based on GPT architecture, while AceGPT \citealp{huang2024acegpt} offers a model designed to adapt to local Arabic culture, specifically tailored to regional nuances. This trend highlights the growing need for multilingual LLMs that perform well in low-resource environments while maintaining competitive performance against more established models.

\paragraph{Vocabulary expansion} 
Vocabulary expansion for large language models (LLMs) has become a crucial area of research, particularly for improving performance in low-resource languages. Traditional methods like Byte Pair Encoding (BPE), while effective at handling out-of-vocabulary (OOV) words, are suboptimal for pretraining larger models, as discussed by Tay et al. (\citealp{bostrom2020byte}), who propose alternative tokenization methods to better capture linguistic nuances. Pham et al. (\citealp{xu2020vocabulary}) advance this by introducing optimal transport-based vocabulary learning, which optimizes the distribution of subword units, enhancing translation tasks, particularly in multilingual and low-resource settings.

Kudo et al. (\citealp{kudo2018subword}) propose subword regularization and offer another avenue for improvement by allowing models to learn from multiple subword segmentation rather than a fixed one, increasing robustness and flexibility. In contexts with limited data, Liu et al. (\citealp{salesky2020optimizing}) have demonstrated that combining subword-based methods with additional pretraining steps significantly improves model performance. These works show that moving beyond traditional vocabulary methods allows for more dynamic and context-aware modeling, enhancing LLMs' scalability and adaptability across diverse linguistic landscapes.

\begin{table*}[htb]
  \centering
  \footnotesize
  \resizebox{\textwidth}{!}{
      \begin{tabular}{llllll}
      \toprule
      \textbf{Aspect}                                   & \textbf{Benchmark}         & \makecell[l]{\textbf{Language} \\\textbf{(+ translation)}} & \textbf{Size} & \textbf{Evaluation Types} & Metrics                                    \\
      \midrule

      \multirow{4}{*}{\makecell[l]{Knowledge Ability}} 
          & RACE~\cite{lai-etal-2017-race} & EN & 4.9K & Multiple-choice Questions & Accuracy \\
          & MMLU~\cite{MMLU21} & EN (+AR) & 14K & Multiple-choice Questions & Accuracy \\
          & ArabicMMLU~\cite{koto2024arabicmmlu} & AR & 14.5K & Multiple-choice Questions & Accuracy \\
          & EXAMS~\cite{EXAMS20} & AR & 0.56K & Multiple-choice Questions & Accuracy \\
      \midrule
 
      \multirow{2}{*}{\makecell[l]{Arabic Cultural \\and Value Alignment}}      
          & ACVA-all~\cite{huang2024acegpt} & AR & 9K & Yes/No binary Questions & F1-score \\
          & ACVA-clean & AR & 2.48K & Yes/No binary Questions & F1-score \\
      \midrule
      \multirow{2}{*}{\makecell[l]{Commonsense \\Reasoning}}   

          & BoolQ~\cite{clark2019boolq} & EN (+AR) & 3.27K & Yes/No binary Questions & Accuracy \\
          & ARC-Challenge~\cite{allenai:arc} & (+AR) & 1.17K & Multiple-choice Questions & Accuracy \\
            
      \bottomrule
      \end{tabular}
  }
  \caption{Overview of Evaluation benchmarks}
  \label{tab:benchmarks}
  \end{table*}

\section{ CEFR Language Proficiency Levels}\label{appdeix:cefr}
Table \ref{tab:cefr} illustrates the vocabulary size that learners are expected to acquire at various stages of second language acquisition.
 The vocabulary size is gradually expanding when humans acquire a second language, as one cannot achieve proficiency in all second-language words at once, as it takes time to digest these words.
\begin{table*}[h]
\centering

\setlength{\tabcolsep}{4pt}
\begin{tabular}{lc|lrr}
\toprule
 \multicolumn{2}{c|}{\textbf{CEFR Level}}  & \textbf{Description} & \textbf{Learning Hours} & \textbf{Vocabulary Size} \\
\midrule
\multirow{2}{*}{Basic User}  & A1 & Beginner Level & 110-130  & 2000 words \\
\cline{2-5}
& A2 & Elementary Level & 150-180  & 3000 words \\
\hline
\multirow{2}{*}{Independent User} & B1& Intermediate Level & 200-230  & 5000 words \\
\cline{2-5}
 & B2 & Upper Intermediate Level & 200-230  & 8000 words \\
\hline
\multirow{2}{*}{Proficient User } & C1 & Advanced Level & 150-200  & 10000 words \\
\cline{2-5}
 & C2 & Mastery Level & 250-300  & 30000 words \\
\bottomrule
\end{tabular}
\caption{CEFR Language Proficiency Levels.}
\label{tab:cefr}
\end{table*}

\section{Cognitive Mechanisms of Vocabulary Acquisition}\label{appdeix:cognitive}
We reviewed relevant literature to confirm the phenomenon of exponential vocabulary expansion in second language acquisition and the cognitive theories that support it. Studies indicate that learners typically begin by mastering a small set of high-frequency vocabulary in the early stages of language learning. As they progress, their vocabulary size grows rapidly. This process can be explained through the following two aspects:

\paragraph{Cognitive Mechanisms of Incremental Learning} In the initial stages, learners build their understanding by repeatedly encountering and using simple foundational words. Research by \cite{krashen1982principles} and \cite{nation2001learning} shows that mastering high-frequency vocabulary is crucial for understanding more complex linguistic structures. These foundational words provide a stable cognitive base, allowing learners to gradually expand their vocabulary \cite{zhang2021effects,nakata2015effects}.

\paragraph{Exponential Vocabulary Growth} Once learners acquire foundational vocabulary, the rate of vocabulary expansion accelerates. Through extensive reading and structured learning strategies such as spaced retrieval practice \cite{leon2016spaced}, learners are able to acquire complex vocabulary in a relatively short period. \cite{coady1996l2} emphasize that extensive reading provides a large amount of language input, enabling learners to incrementally encounter and absorb more advanced vocabulary.

\section{ Arabic data distribution}\label{appdeix:data distribution}
\jianqing{Table \ref{tab:pretrain data} show the Arabic dataset primarily draws from several key sources, with the largest contribution coming from the Common Crawl (filtered) dataset, which accounts for 55.5\% of the total data. Other significant sources include WebText, which contributes 26.7\%, and Books+Newspapers, providing 8.9\% with 2.5 billion tokens. Additionally, Wikipedia is divided into two parts, contributing 3.76\% and 5.14\%. These diverse sources collectively form the foundation for training the Arabic model.}

\begin{table*}[ht]
\centering
\label{tab:pretrain data}
\begin{tabular}{llr}
\toprule
\textbf{Dataset} & \textbf{ \# tokens} & \textbf{Weight in training mix} \\
\midrule
Common Crawl (filtered) & 101.3 billion & 55.5\%  \\
WebText & 10.62 billion & 26.7\%  \\
Books+Newspapers & 2.5 billion & 8.9\%  \\
Wikipedia1 & 0.36 billion & 3.76\%  \\
Wikipedia2 & 0.51 billion & 5.14\% \\
\bottomrule
\end{tabular}
\caption{Arabic data distribution and elapsed epochs} 
\label{tab:pretrain data}
\end{table*}

\section{Data mixture}\label{appdeix:data mixture}
\label{app:mixture}

Table \ref{tab:data_ratio} shows the data distribution across the pre-training stages is carefully adjusted, with the proportions of Arabic and English data determined using a cosine annealing schedule. Initially, the Arabic data constitutes 30\% of the total, while English data makes up 65\% and math \& coding data consistently accounts for 5\%. As the training progresses and new subwords are added, the proportion of Arabic data increases steadily, reaching 90\% by the final stage. Concurrently, the English data proportion decreases to 5\%, while the math \& coding data remains constant at 5\% throughout all stages. This dynamic adjustment ensures that the model effectively balances the learning of Arabic and English content, with a strong emphasis on Arabic in the later stages.

\begin{table*}[ht]
\small
\begin{center}
\begin{tabular}{ccccr}
\toprule
\textbf{Stage} & \textbf{New subwords added} & \textbf{Arabic data } & \textbf{English data } & \textbf{math \& coding data } \\
\midrule
1 & 0 & 30.00\% & 65.00\% &5.00\% \\
2 & 1 & 30.33\% & 64.47\% &5.00\% \\
3 & 2 & 31.31\% & 63.69\% &5.00\% \\
4 & 4 & 32.94\% & 62.06\% &5.00\% \\
5 & 8 & 35.19\% & 59.81\% &5.00\% \\
6 & 16 & 38.04\% & 56.96\% &5.00\% \\
7 & 32 & 41.46\% & 53.54\% &5.00\% \\
8 & 64 & 45.41\% & 49.59\% &5.00\% \\
9 & 128 & 49.85\% & 45.15\% &5.00\% \\
10 & 256 & 54.73\% & 40.27\% &5.00\% \\
11 & 512 & 60.00\% & 35.00\% &5.00\% \\
12 & 1024 & 65.60\% & 29.40\% &5.00\% \\
13 & 2048 & 71.46\% & 23.54\% &5.00\% \\
14 & 4196 & 77.53\% & 17.47\% &5.00\% \\
15 & 8192 & 83.73\% & 11.27\% &5.00\% \\
16 & 12800 & 90.00\% & 5.00\% &5.00\% \\
\bottomrule
\end{tabular}
\caption{Detailed distribution of Arabic, English and math \& coding data across each pre-training stage.}
\label{tab:data_ratio}
\end{center}
\end{table*}

\section{Comparison of compression ratio and OOV changes at different stages between exponential and uniform expansion}\label{appdeix:compress ratio and oov rate}
Table \ref{tab:oov rate comparison} illustrates the trends in compression ratio and OOV (Out-Of-Vocabulary) ratio as vocabulary size is incrementally expanded using both Exponential and Uniform methods. In the case of **Exponential Vocabulary Expansion**, both the compression ratio and OOV ratio change gradually, ensuring a more balanced progression as new subwords are added. This gradual change is beneficial for maintaining stability during model training, as it allows the system to adjust incrementally to the growing vocabulary. 

\begin{table*}[h]
\centering
\resizebox{\textwidth}{!}{
\begin{tabular}{cccccc}
\toprule
\textbf{Add Subword Size} & \textbf{Compress Ratio (Exponential)} & \textbf{OOV Ratio (Exponential)} & \textbf{Add Subword Size} & \textbf{Compress Ratio (Uniform)} & \textbf{OOV Ratio (Uniform)} \\
\midrule
0 & 0.90 & 0.000 & 0 & 0.90 & 0.000 \\
1 & 0.88 & 0.017 & 853 & 0.45 & 0.669 \\
2 & 0.87 & 0.018 & 1736 & 0.40 & 0.116 \\
4 & 0.85 & 0.022 & 2559 & 0.37 & 0.068 \\
8 & 0.82 & 0.038 & 3412 & 0.35 & 0.049 \\
16 & 0.77 & 0.061 & 4265 & 0.34 & 0.039 \\
32 & 0.72 & 0.076 & 5118 & 0.33 & 0.031 \\
64 & 0.65 & 0.094 & 5971 & 0.32 & 0.026 \\
128 & 0.60 & 0.093 & 6824 & 0.31 & 0.021 \\
256 & 0.54 & 0.105 & 7677 & 0.31 & 0.019 \\
512 & 0.48 & 0.116 & 8530 & 0.30 & 0.017 \\
1024 & 0.43 & 0.110 & 9383 & 0.30 & 0.015 \\
2048 & 0.39 & 0.118 & 10236 & 0.30 & 0.013 \\
4096 & 0.34 & 0.120 & 11089 & 0.29 & 0.012 \\
8192 & 0.31 & 0.116 & 11942 & 0.29 & 0.011 \\
12800 & 0.28 & 0.070 & 12800 & 0.28 & 0.010 \\
\bottomrule
\end{tabular}}
\caption{Comparison of Exponential and Uniform Vocabulary Expansion Methods}
\label{tab:oov rate comparison}
\end{table*}

\section{Evaluation of models in English MMLU dataset}\label{enmmlu}

In the evaluation of English MMLU performance, \model models, both 7B and 13B, consistently outperform their counterparts across most categories in both few-shot and zero-shot settings (shown in Table \ref{tab:mmlu-result}). Particularly, \modelLarge achieves the highest average score of 62.89 in zero-shot tasks, demonstrating its superior capability in generalization and task adaptability. 

\begin{table*}[H]
\setlength{\tabcolsep}{2pt}
\centering
\footnotesize
\resizebox{\textwidth}{!}{
\begin{tabular}{l|llll|l|llll|l|l}
\toprule
                & \multicolumn{5}{c}{few shot English MMLU}& \multicolumn{5}{c}{zero shot English MMLU} &\textbf{Total}  \\
Model           &  STEM & \makecell[c]{Human-\\ities} &\makecell[c]{Social \\ Sciences}   & Others &Avg.& STEM & \makecell[c]{Social \\ Sciences} &  \makecell[c]{Human-\\ities}    & Other  & Avg. &\textbf{Avg.}\\
\midrule
LLaMA2-7B   &\textbf{40.00} &\textbf{51.95} &52.42 &\textbf{50.89} &\textbf{48.81} & 31.49& 31.26 &38.35 &38.80 &34.97&41.89\\
AceGPT-7B  & 36.09 &46.33 &49.19 &46.23 &44.46 & 33.91& 43.85 &49.47 &45.38 &43.15&43.81\\
\textbf{\modelSmall}  & 38.44 &49.62& \textbf{53.32}& 50.61& 48.00&  \textbf{45.49}& \textbf{63.55}& \textbf{66.05}& \textbf{59.25}& \textbf{58.59}&\textbf{53.30}\\
\midrule
LLaMA2-13B  & 47.28& \textbf{63.55}& {64.33}& {57.97}& \textbf{58.28}&-&-&-&-&-&- \\
Jais-13B &27.14& 14.38& 45.64& 41.13& 32.11&44.33 &55.14 &61.39& 56.06& 54.23&43.17 \\
AceGPT-13B  & 46.66 &61.39& 63.37& 56.12& 56.88&  39.88& 52.18& 58.51& 49.61& 50.04&53.46\\
\textbf{\modelLarge} & \textbf{47.31} &62.47& \textbf{64.77}& \textbf{58.14}& 58.17& \textbf{ 51.48}& \textbf{66.71} &\textbf{71.65}& \textbf{61.72}& \textbf{62.89}&\textbf{60.53}\\
\midrule
Jais-30B   & 27.42& 14.60& 45.84& 41.43& 32.32&  39.23& 44.51& 52.96& 50.91& 46.90&39.61\\
GPT-3.5 Turbo  &58.39& 72.12& 78.02& 69.95& 69.62&-&-&-&-&-&-\\
\bottomrule
\end{tabular}
}
\caption{Evaluation of models in English MMLU dataset: few-shot on base model and zero-shot on chat model. Under the zero-shot setting, LLaMA2-13B model does not follow instructions for unknow reason.}
\label{tab:mmlu-result}
\end{table*}

\section{ALAN examples}\label{alan}
\label{app:alan_examples}
We provide concrete examples of ALAN below. Note that we translate examples into English using {\tt GPT-3.5-Turbo}. In practice, our data is in Arabic.

\subsection{Topics}
A set of 30 topics, randomly chosen, is listed below:

\texttt{\scriptsize
"Arabic Language and Literature"
"Mathematics"
"Islamic Studies"
"Middle Eastern History and Politics"
"Computer science"
"Economics"
"Healthcare industry"
"Social work"
"Business"
"Geography"
"Mining"
"Chemical Engineering"
"Languages and Literature"
"Materials Science and Engineering"
"Transport industry"
"Chemistry"
"Food industry"
"Systems science"
"Astronomy"
"Cultural industry"
"Energy industry"
"Radiology"
"Pediatrics"
"Dentistry"
"Civil Engineering"
"Aerospace industry"
"Public administration"
"Infectious disease"
"Public policy"
"Environmental studies and forestry"
}

\subsection{Subjects}
A set of 30 subjects, randomly chosen, is listed below:

\texttt{\scriptsize
"Hypersonic and High-Speed Flows"
"Mental Health Nursing"
"Mechanical Systems and Energy Efficiency"
"Obstetrics and Gynecological Nursing"
"Immunology"
"Interdisciplinary Geriatric Care"
"Signal Processing"
"Geography research methods and techniques"
"Public Administration and Management"
"An introduction to space exploration"
"Environmental and Safety Management"
"Social and Ethical Aspects of Agriculture"
"Folk and Cultural Dance"
"Power System Protection and Control"
"Collage and Mixed Media"
"Advanced Game Theory"
"Pediatric Critical Care"
"Transport Modeling and Forecasting"
"Foundations of Mathematics"
"Carbon Capture, Storage, and Utilization"
"Customer Service and Relationship Management"
"Introduction to Probability"
"Virtual Reality and Augmented Reality"
"Reservoir Management and Enhanced Oil Recovery"
"Safety and Standards in Industrial Robotics"
"Social Work with LGBTQ+ populations"
"Nutritional Science"
"Advanced Gynaecology Courses"
"Bioinformatics and Computational Chemistry"
"Reusable Launch Vehicle Technology"
}

\subsection{A syllabus with specific knowledge points}
We provide an example syllabus with specific knowledge points as below.

\texttt{\scriptsize
Subject title: Hypersonic and High-Speed Flows \\
  Lecture title: Introduction to Hypersonic Flows \\
    Knowledge points: \\
    - Definition of hypersonic flows \\
    - Mach number \\
    - Key characteristics of hypersonic flows \\
  Lecture title: Fundamentals of Shock Waves \\
    Knowledge points: \\
    - Definition of shock waves \\
    - Formation of shock waves \\
    - Types of shock waves \\
  Lecture title: High-Temperature Gas Dynamics \\
    Knowledge points: \\
    - Definition of high-temperature gas dynamics \\
    - Behavior of high-temperature gases \\
    - Effects of high-temperature gases on materials \\
  Lecture title: Principles of Rarefied Gas Dynamics \\
    Knowledge points: \\
    - Definition of rarefied gas dynamics \\
    - The continuum hypothesis \\
    - Governing equations \\
  Lecture title: High-Speed Flow Over Bodies \\
    Knowledge points: \\
    - High-speed flow characteristics \\
    - Impact on the body \\
    - Aerodynamic heating \\
  Lecture title: Hypersonic Vehicle Configurations \\
    Knowledge points: \\
    - Types of hypersonic vehicles \\
    - Vehicle configurations \\
    - Advantages and limitations of each configuration \\
  Lecture title: Aerothermodynamics of Hypersonic Flows \\
    Knowledge points: \\
    - Definition of aerothermodynamics \\
    - Aerothermodynamics in hypersonic flows \\
    - Heat transfer in hypersonic flows \\
  Lecture title: Hypersonic Flow Control \\
    Knowledge points: \\
    - Importance of flow control \\
    - Methods of hypersonic flow control \\
    - Challenges in hypersonic flow control \\
  Lecture title: Hypersonic Propulsion Systems \\
    Knowledge points: \\
    - Types of hypersonic propulsion systems \\
    - Working principles \\
    - Advantages and disadvantages \\
  Lecture title: Future Trends in Hypersonic and High-Speed Flows \\
    Knowledge points: \\
    - Current research in the field \\
    - Potential future trends \\
    - Challenges and opportunities
}

\subsection{Synthetic QA data}
We provide a synthetic QA example using knowledge points generated by {\tt GPT-3.5-Turbo}.

\texttt{\scriptsize
Subject title: \\
Computer Vision for Industrial Robotics \\
 \\
Lecture title: \\
Stereo Vision and 3D Reconstruction \\
 \\
Knowledge points: \\
- Principles of stereo vision \\
- Stereo camera calibration \\
- Depth estimation and 3D reconstruction \\
- Point cloud processing \\
 \\
Synthetic question: \\
In stereo vision, the process of determining the depth of objects in a scene is known as: \\
A. Image rectification \\
B. Disparity mapping \\
C. Camera calibration \\
D. Point cloud processing \\
 \\
Synthetic solution to the question: \\
B \\
 \\
Explanation: \\
 \\
The correct answer is B. Disparity mapping. In stereo vision, the depth of objects in a scene is determined by calculating the disparity between corresponding points in the left and right images. Disparity mapping involves finding the pixel-level differences between the two images to estimate the depth information.
}

\section{Instruction-following test}\label{sec:instrucion-following}
We evaluated the models' instruction-following capabilities using the Arabic versions of Vicuna-80~\cite{vicuna2023}, translated by GPT-4 and refined by native speakers. Following the methodology in~\cite{vicuna2023}, GPT-4 was used as the evaluator, assigning scores to each model's performance compared to GPT-3.5 Turbo, with a temperature setting of 0.2. For each question, GPT-4 independently scored the responses from both the evaluated model and GPT-3.5 Turbo. The average performance ratio of the evaluated model was calculated by dividing its overall score by that of GPT-3.5 Turbo. Results in Table~\ref{tab:vicuna-80} indicate that \model models outperform their counterparts in Arabic Vicuna-80. Notably, \model-7B exceeds Jais-13B by approximately 17\%, despite having a smaller model size.

\begin{table}[h]
\centering
\setlength{\tabcolsep}{3pt}
\begin{tabular}{lr}
\toprule
Model         & Ratio of GPT-3.5 \\
\midrule
Jais-13B&	75.40\% \\
Llama-7B&	78.99\%  \\
\model-7B&	92.71\% \\
\bottomrule
\end{tabular}
\caption{Performance ratio of GPT-3.5 Turbo in Arabic Vicuna-80.}
\label{tab:vicuna-80}
\end{table}

\section{Details of Ablation Study}
\label{appendix: details of ablation}

\begin{figure}[h]
\centering
\includegraphics[width=0.5\textwidth]{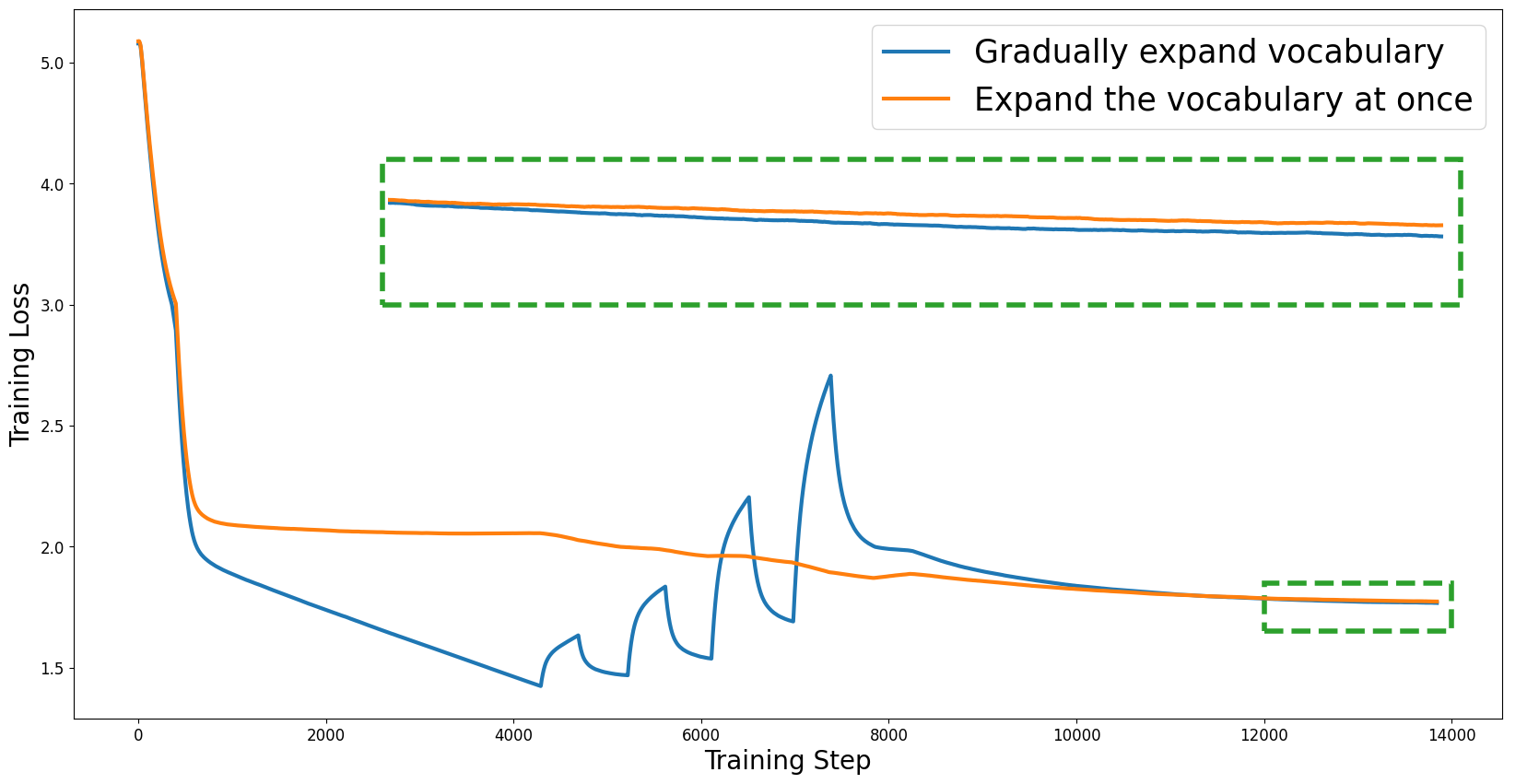}
\caption{Loss curve of TinyLLaMa with sliding window average}
\label{fig:loss-1B}
\end{figure}

\subsection{Experiment Settings: } 
We undertook continuous pre-training on a 1B-parameter TinyLLaMA model~\cite{zhang2024tinyllama}, which is derived from the LLaMA architecture and was initially trained on an English corpus comprising 3 trillion tokens. The pre-training regimen was segmented into five distinct stages, during which 0, 16, 64, 256, and 1024 Arabic subwords were progressively added to the vocabulary. Each stage allocated a different volume of data, totaling 80 billion tokens, with the proportion of Arabic to English data gradually shifting from 0:10 to 9:1. In a parallel experiment, we introduced 1024 subwords to the vocabulary in a single step, maintaining the same total token count and data distribution as in the phased approach. Both experiments adhered to an identical learning rate strategy, reinstating a cosine learning rate scheduler at the onset of each stage, starting with an initial rate of 1e-5 and tapering to 2e-6, with the initial 5 billion tokens of each stage designated for warm-up. Utilizing 192 GPUs, the experiments were conducted with a batch size of 3072. 

\begin{table*}[htb]
\centering
\setlength{\tabcolsep}{3pt}
\begin{tabular}{l|lllll|l}
\toprule
Model         & STEM & \makecell{Social\\Sciences} & Humanities & \makecell{Arabic\\Language} & Other & Avg  \\
\midrule

Expand vocab at once      	&28.6	&26.7	&\textbf{28.1}&	24.4	&30.1& 27.0 \\
\textbf{Gradually expand vocab (ours)  }     	&\textbf{29.8}&	\textbf{27.1}&	27.2&	\textbf{24.6}&	\textbf{31.4}&\textbf{27.3}  \\
\bottomrule
\end{tabular}
\caption{Zero-shot evaluation for TinyLLaMA in ArabicMMLU~\cite{koto2024arabicmmlu} with option logit probabiltiy}
\label{tab:tinyllama-mbzuai}
\end{table*}

\subsection{Progressive Vocabulary Expansion Pre-training}
\juhao{The results shown in Figure~\ref{fig:loss-1B} demonstrate that the strategy of progressively expanding the vocabulary, which applies a sliding window average technique, yields a reduced final loss. 
Furthermore, as evidenced in Table \ref{tab:tinyllama-mbzuai}, within the ArabicMMLU dataset, the approach of incrementally introducing new vocabulary items consistently outperforms the method of a one-time vocabulary expansion. This pattern underscores the effectiveness of gradual vocabulary enhancement in optimizing model performance.}

\end{document}